\documentclass[11pt,a4paper]{article}
\usepackage[hyperref]{acl2020}
\usepackage{times}
\usepackage{latexsym}
\usepackage{xspace}
\usepackage{url}
\usepackage{mathtools}
\usepackage{amssymb}
\usepackage{arabtex}
\usepackage{footnote}
\usepackage{tablefootnote}
\usepackage{utf8}
\usepackage{graphicx}
\usepackage{color, colortbl}
\usepackage{xstring}
\usepackage{comment}
\usepackage{array,booktabs,makecell}

\usepackage{microtype}

\usepackage[english]{babel}
\usepackage{fancyhdr}
\usepackage{lastpage}
 
\aclfinalcopy 

\title{Sentence-Level BERT and Multi-Task Learning of Age and Gender in Social Media}

\author{Muhammad Abdul-Mageed$^1$  Chiyu Zhang$^1$  Arun Rajendran$^1$  \\ \textbf{AbdelRahim Elmadany$^1$ Michael Przystupa$^1$ Lyle Ungar$^2$}  \\ 
  $^1$Natural Language Processing Lab,  University of British Columbia \\
  $^2$Computer and Information Science, University of Pennsylvania \\
  {$^1$\tt muhammad.mageed@ubc.ca $^2$\tt  ungar@cis.upenn.edu}\\}

\begin{document}
\maketitle
\begin{abstract}
Social media currently provide a window on our lives, making it possible to learn how people from different places, with different backgrounds, ages, and genders use language. In this work we exploit a newly-created Arabic dataset with ground truth \textit{age} and \textit{gender} labels to learn these attributes both individually and in a multi-task setting \textit{at the sentence level}. Our models are based on variations of deep bidirectional neural networks. More specifically, we build models with gated recurrent units and bidirectional encoder representations from transformers (BERT). We show the utility of multi-task learning (MTL) on the two tasks and identify task-specific attention as a superior choice in this context. We also find that a single-task BERT model outperform our best MTL models on the two tasks. We report tweet-level accuracy of 51.43\% for the age task (three-way) and 65.30\% on the gender task (binary), both of which outperforms our baselines with a large margin. Our models are language-agnostic, and so can be applied to other languages. 
\end{abstract}

\setcode{utf8}
\setarab
\novocalize
\section{Introduction}

Wide use of social media has made it possible to collect large amounts of data from people belonging to various types of backgrounds. These data can then be used for developing natural language processing models capable of predicting user attributes such as location, gender, age, and personality type. While there has been multiple works on learning these tasks, most of them focus on single-task learning rather than trying to learn more than one task simultaneously. This is sub-optimal and different from how humans learn~\cite{caruana1997multitask}. 
\begin{figure}[h]
\begin{centering}
\frame{\includegraphics[width=\linewidth]{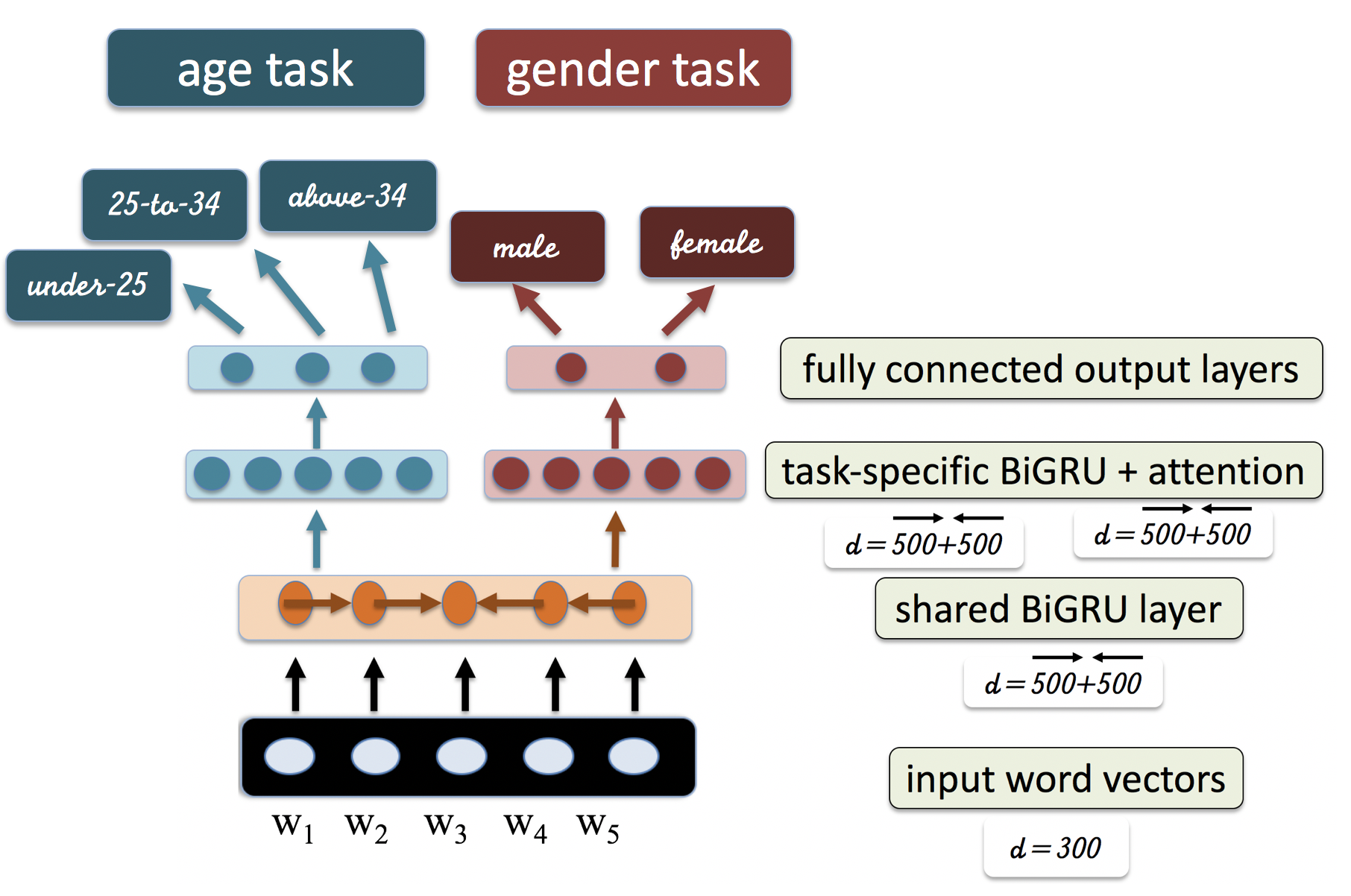}}
  \caption{Illustration of our MTL network with shared hiddent layers and a task-specific attention layer.}
  \label{fig:mtl_tas-spec-attn}
  \end{centering}
\end{figure}

In this work, we investigate the utility of multi-task learning (MTL), where more than one task is learned simultaneously in a single model with an overall loss that is the sum of each independent task's loss. We focus on learning the two tasks of age and gender at the \textit{sentence level}, exploring a deep neural network paradigm where the model's hidden layers are shared in the MTL setting with (1) a shared attention layer but also (2) with task-specific attention layers. An illustration of our MTL network with task-specific attention is presented in Figure~\ref{fig:mtl_tas-spec-attn}.

In a related vein, recent progress in representation learning of natural language has resulted in a proliferation of methods. These include methods aimed at improving input to the neural network by learning better embeddings either exploiting parallel data in machine translation tasks (as in CoVe~\cite{mccann2017learned}) or by enhancing these embedding by exploiting bidirectional language models (as in ELMo~\cite{peters2018deep}). Other approaches such as the Transformer model~\cite{vaswani2017attention}, which dispenses entirely with both the recurrence and convolution approaches dominant in neural networks, and its extensions such as the work of ~\cite{radford2018improving} and ~\cite{devlin2018bert} that leverage Transformers for pre-training. These methods have hardly been used in Arabic NLP and hence it remains unclear what their utilities are and how it is that they can be best deployed for the language. Toward bridging this gap, we explore the benefit of using a specific type of Transformers based on bidirectional, masked language models (BERT) ~\cite{devlin2018bert}.

Finally, we note that in spite of the Arabic world being a large region with a huge population (~300M) and a wide use of social media, tools for understanding attributes of Arabic users online are still lacking. In spite of the important and worthy privacy concerns related to user online behavior, such tools can be valuable for making discoveries about whole communities and as means to facilitate humanitarian (as in disaster), educational (as in targeted mediated pedagogy), and safety causes (as in detecting hate speech and extremist ideology), to list a few. Our models offer a step forward toward building tools for online Arabic user modeling.

Overall, we make the following contributions: (1) We develop novel models targeting the two tasks of age and gender; (2) we investigate and establish the utility of multi-task learning on the these two tasks, showing the advantage of task-specific attention (vs. shared attention, see Section~\ref{subsec:mtl}); and (3) we carry out BERT-based modeling on the two tasks. The rest of the paper is organized as follows: In Section~\ref{sec:data} we describe the dataset. In Section~\ref{sec:methods} we introduce our various methods. We present our experimental setup and our models in Section~\ref{sec:exps}, and related works in Section~\ref{sec:rel}. We conclude in Section~\ref{sec:conc}.

\section{Data}\label{sec:data}
\begin{figure}[h]
\begin{centering}
 \frame{\includegraphics[width=\linewidth]{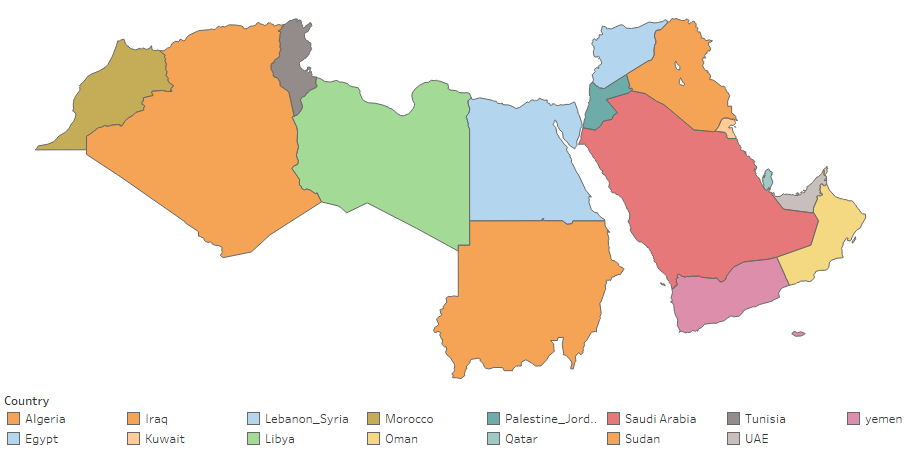}}
  \caption{Arabic countries covered in the dataset.}
  \label{fig:ara_w_wiki}
  \end{centering}
\end{figure}
\begin{table*}[h]
\centering
\resizebox{\textwidth}{!}{%
\begin{tabular}{llllllll}
\hline
\textbf{Data split} & \multicolumn{2}{c}{\textbf{Under 25}} & \multicolumn{2}{c}{\textbf{25 until 34}} & \multicolumn{2}{c}{\textbf{35 and up}} & \textbf{\#tweets} \\
\textbf{} & \textbf{Female} & \textbf{Male} & \textbf{Female} & \textbf{Male} & \textbf{Female} & \textbf{Male} & \textbf{} \\ \hline
\textbf{TRAIN} & 215,950 & 213,249 & 207,184 & 248,769 & 174,511 & 226,132 & 1,285,795 \\
\textbf{DEV} & 27,076 & 26,551 & 25,750 & 31,111 & 21,942 & 28,294 & 160,724 \\
\textbf{TEST} & 26,878 & 26,422 & 25,905 & 31,211 & 21,991 & 28,318 & 160,725 \\ \hline
\textbf{ALL} & 269,904 & 266,222 & 258,839 & 311,091 & 218,444 & 282,744 & 1,607,244 \\\hline
\end{tabular}%
}
\caption{Distribution of age and gender classes in our data splits}
\label{tab:classes}
\end{table*}

We make use of Arap-Tweet~\cite{zaghouani2018arap}, which we will refer to as \textit{Arab-Tweet}. Arab-tweet is a dataset of tweets covering 11 Arabic regions from 17 different countries.~\footnote{Please note~\cite{zaghouani2018arap} mention the dataset covers 16 countries, but in the distribution we received we found the data cover 17 countries.} For each region, data from 100 Twitter users were crawled. Users needed to have posted at least 2,000 and were selected based on an initial list of seed words characteristic of each region. The seed list included words such as <برشة> /barsha/ ‘many’ for Tunisian Arabic and  <وايد> /wayed/ ‘many’ for Gulf Arabic.~\cite{zaghouani2018arap} employed human annotators to verify that users do belong to each respective region. Annotators also assigned gender labels from the set \textit{{male, female}} and age group labels from the set \textit{{under-25, 25-to34, above-35}} at the user-level, which in turn is assigned at tweet level. Tweets with less than 3 words and re-tweets were removed. Refer to~\cite{zaghouani2018arap} for details about how annotation was carried out. We provide a description of the data in Table~\ref{tab:classes}. (Table~\ref{tab:classes} also provides class breakdown across our splits, as we explain in more detail in Section~\ref{subsec:splits}).~\footnote{We provide randomly sampled examples from the various age and gender categories in the supplementary material.} We note that~\cite{zaghouani2018arap} do not report classification models exploiting the data.

\section{Methods}\label{sec:methods}
\subsection{GRU}

For comparison, we employ gated recurrent units (GRU)~\cite{cho2014learning}, a simplification of long-short term memory networks (LSTMs)~\cite{hochreiter1997long}. Like LSTMs, GRUs can address the problem of long-range dependencies in recurrent neural networks. Unlike an LSTM, a GRU does not have a memory cell but makes use of an \textit{update gate} $\textbf{\textit{z}}^{(t)}$ and a \textit{reset gate} $\textbf{\textit{r}}^{(t)}$. The activation of GRU at time step $t$ is a linear interpolation of the previous activation \textit{hidden state} $\textbf{\textit{h}}^{(t-1)}$ and the candidate activation \textit{hidden state} \begin{math}  \textbf{\textit{$\widetilde{h}$}}^{(t)} \end{math}. The \textit{update state} $\text
{\textit{z}}^{(t)}$ decides how much the unit updates its content. Essentially, the forget gate of an LSTM and an update gate of a GRU each acts as a manager of information flow where they allow the network to update its state by adding content from the current cell to that of a previous cell, rather than overwriting previous cell content (as is the case in vanilla RNNs).
\subsection{Multi-Task Learning}
In addition to attempting to learn each of the two tasks independently, we explore the utility of learning them together in a \textit{multi-task learning} (MTL) framework. The main intuition behind MTL is that most real-world tasks are multi-task problems that humans learn together. Hence, when we treat problems as single-tasks, we might be sacrificing performance~\cite{caruana1993}. For related tasks, MTL helps achieve inductive transfer between these tasks by leveraging additional sources of information (from other tasks) to improve performance on a given task (current task). ~\cite{caruana1997multitask}. MTL also helps regularize models. The framework has been successfully employed to several NLP tasks such as machine translation and syntactic parsing~\cite{luong2015multi} and syntactic chunking, POS tagging, and CCG supertagging~\cite{sogaard2016deep}.

In the single task scenario, an independent network is designed for each task. Each network is trained in isolation, with its own backpropagation. For example, Figure ~\ref{fig:stl_net} is an illustration of our network for learning gender and a similar, independent network is devoted for learning age. In MTL, these tasks would be learned together in a single network with 2 different outputs, one for each task. The MTL network would have one or more hidden layers that are shared between all the tasks. Backpropagation is then applied in parallel on the 2 outputs.Figure~\ref{fig:mtl_tas-spec-attn} presented earlier is an illustration of an MTL network for learning age and gender, with 1 shared hidden layer. 

In our current work, each of the two independent networks has its own loss function. Formally, let $\{x^{(i)}, y^{(i)}\}_{i=1}^{N}$ denote the training data, where $x^{(i)} \in R^d$ is the input vector, $y^{(i)} \in \{1, \dots, C^{(y)}\}$ is the class label of the \textit{i}-th data point, \textit{d} is the dimensionality of the input vector, $C^{(y)}$ is the number of classes, and \textit{N} is the number of data points. The cross-entropy function is calculated as:

\begin{equation}
 \mathcal{L}(\theta)\\= - \sum_{i} log P\left(y^{(i)}|x^{(i)}; \theta \right) \\ 
\end{equation}



For gender, we calculate binary cross-entropy with $y^{(i)} \in \{0=male, 1=female\}$, and in age we compute a categorical cross-entropy as $y^{(i)} \in \{0=under-25, 1=25-to-34, 2=above-34\}$. Our MTL loss can then be computed as:

\begin{equation}
 \mathcal{L}(\theta_{MTL}) =  \left(\mathcal{L}(\theta_{age}) + \mathcal{L}(\theta_{gender}) \right) / 2 \\ 
\end{equation}


We now introduce the Transformer~\cite{vaswani2017attention}, on which our attention mechanism is based. The Transformer is also the backbone of the BERT model~\cite{devlin2018bert} that we employ as we will explain in Section~\ref{subsec:bert}.

\subsection{Transformer}
~\citet{vaswani2017attention} introduced the Transformer, which is based solely on attention, thus allowing for parallelizing the network and dispensing of recurrence and convolution. Similar to sequence transduction models such as ~\citet{bahdanau2014neural,cho2014learning,Sutskever2014SequenceTS}, the Transformer is an encoder-decoder architecture. The encoder takes a sequence of symbol representations $x^{(i)} \dots x^{(n)}$, maps them into a sequence of continuous representations $z^{(i)} \dots x^{(n)}$ that are then used by the decoder to generate an output sequence $y^{(i)} \dots y^{(n)}$, one symbol at a time. This is performed using \textit{self-attention}, where different positions of a single sequence are related to one another. The Transformer employs an attention mechanism based on a function that operates on \textit{queries}, \textit{keys}, and \textit{values}. The attention function maps a query and a set of key-value pairs to an output, where the output is a weighted sum of the values. For each value, a weight is computed as a compatibility function of the query with the corresponding key. 


We implement this attention method in our BiGRU models as an attention layer. As such, we still use a BiGRU architecture that is a variation of RNNs but that has the Transformer's self-attention mechanism. In our BiGRUs, the query is the hidden state vector from the BiGRU and both the key and value are the same output vector of the network. The query, key, and value have the same dimension $d_k=500$. This particular version of attention is a scaled dot product of queries and keys (each of $d_k$) that is scaled by a factor of $\frac{1}{\sqrt{d_k}}$ on which a softmax is applied to acquire the weights on the values. The scaled dot product attention is computed as as a set of queries, keys, and values in three matrices Q, K, and V, respectively, follows:

\begin{equation}
 Attention \left(\textit{Q, K, V}\right) = softmax \left( \frac{QK^T}{\sqrt{d_k}}\right)V  \\ 
\end{equation}

Encoder of the Transformer in~\cite{vaswani2017attention} has 6 attention layers, each of which is composed of two sub-layers. The first sub-layer is a \textit{multi-head attention} where, rather than performing a single attention function with queries, keys, and values, these are are projected \textit{h} times into linear, learned projections and ultimately concatenated. The second sub-layer is a simple, fully-connected feed-forward network (FFN), with two linear layers and a ReLU activation function in-between, that is applied to each position separately and identically. Decoder of the Transformer also employs 6 identical layers, similar to the encoder, yet with an extra/third sub-layer that performs multi-head attention over the encoder stack. More information about the Transformer can be found in ~\citet{vaswani2017attention}. In addition to Transformer self-attention, we also experiment with \textit{supervised attention} based on mutual information, which we now introduce.

\subsection{Supervised Attention.}\label{subsec:suprvsdattn}

We propose a novel method of supervised attention that uses implicit vocabulary list obtained from \textit{mututal information (MI)}. MI is a measure of the similarity between two labels (e.g., ``male" and ``female" in the case of \textit{gender}) of the same N objects (discrete features, or words in our case). Where 
$|U^i|$ is the number of the samples in class $U^i$ and $|V^j|$ is the number of the samples in class $V^j$, the MI between clusterings 
$U$ and  $V$ is given (in set cardinality formulation) by:
~\footnote{We use scikit-learn to acquire MI over words. For more information, see: \url{https://scikit-learn.org/stable/modules/clustering.html\#mutual-info-score}.}

 \begin{equation}
 MI(U,V)=\sum_{i=1}^{|U|} \sum_{j=1}^{|V|} \frac{|U^{i}\cap V^{j}|}{N}
        \log\frac{N|U^{i} \cap V^{j}|}{|U^{i}||V^{j}|}  
 \end{equation}

 We take the top 500 words based on mutual information score for each label and we use this list of words to perform the supervised attention. We employ the supervised attention for each of the labels using their respective lists. The gold attention weights are calculated based on the presence of the words from the list. For example, if we have four words in the sentence from the list, then our gold attention weights vector will have 0.25 (1/4) at the corresponding indexes. In absence of all words from the gold list, we follow ~\cite{liu2017attention} in using the regular attention mechanism, thus allowing our models to perform our multi-head attention. As mentioned earlier, we also use BERT~\cite{devlin2018bert} to learn age and gender independently. We now introduce BERT.

\subsection{BERT}\label{subsec:bert}
BERT~\cite{devlin2018bert} stands for \textbf{B}idirectional \textbf{E}ncoder \textbf{R}epresentations from \textbf{T}ransformers. BERT is an approach for pre-training language representations. BERT involves two unsupervised learning tasks, (1) \textit{masked language models (Masked LM)} and (2) \textit{next sentence prediction}. Since BERT uses bidirectional conditioning, a given percentage of random input tokens are masked and the model attempts to predict these masked tokens. This is the Masked LM task, where masked tokens are simply replaced by a string \texttt{[MASK]} .~\cite{devlin2018bert} mask 15\% of the tokens (the authors use WordPieces) and feed the final hidden vectors of these masked tokens to an output softmax over the vocabulary. The next sentence prediction task of BERT is also straightforward. The authors simply cast the task as binary classification. For a given sentence \texttt{S}, two sentences \texttt{A} and \texttt{B} are generated where \texttt{A} (positive class) is an actual sentence from the corpus and \texttt{B} is a randomly chosen sentence (negative class). Once trained on an unlabeled dataset, BERT can then be fine-tuned with supervised data for a downstream task (e.g., text classification, question answering). We now introduce our experiments.

\section{Experiments}\label{sec:exps}
We detail how we split and pre-process our data and list our baselines and evaluation metrics here. We then present (i) our single-task models (Section ~\ref{subsec:stl}), (ii) multi-task models (Section ~\ref{subsec:mtl}), and (iii) the BERT model (Section ~\ref{subsec:bert_results}). 

\subsection{Data Splits, Baselines, and Evaluation}\label{subsec:splits}
We shuffle the dataset and split it into 80\% training (TRAIN), 10\% development (DEV), and 10\% test (TEST). The distribution of classes in our splits is in Table~\ref{tab:classes}. For pre-processing, we reduce 2 or more consecutive repetitions of the same character into only 2 and remove diacritics. We have two baselines, (1) (\textit{maj-base}), is the majority class in our TRAIN and (2) a small unidirectional GRU (\textit{small-GRU}) with a single 500-units hidden layer. We train the small GRU with the same batch size = 8 and dropout =0.5 as our bigger BiGRUs (see Section~\ref{subsec:stl}), but with no attention. We report our results in \textit{accuracy}. 

\subsection{Single-Task BiGRUs}\label{subsec:stl}
\begin{figure}[h]
\begin{centering}
\frame{\includegraphics[width=\linewidth]{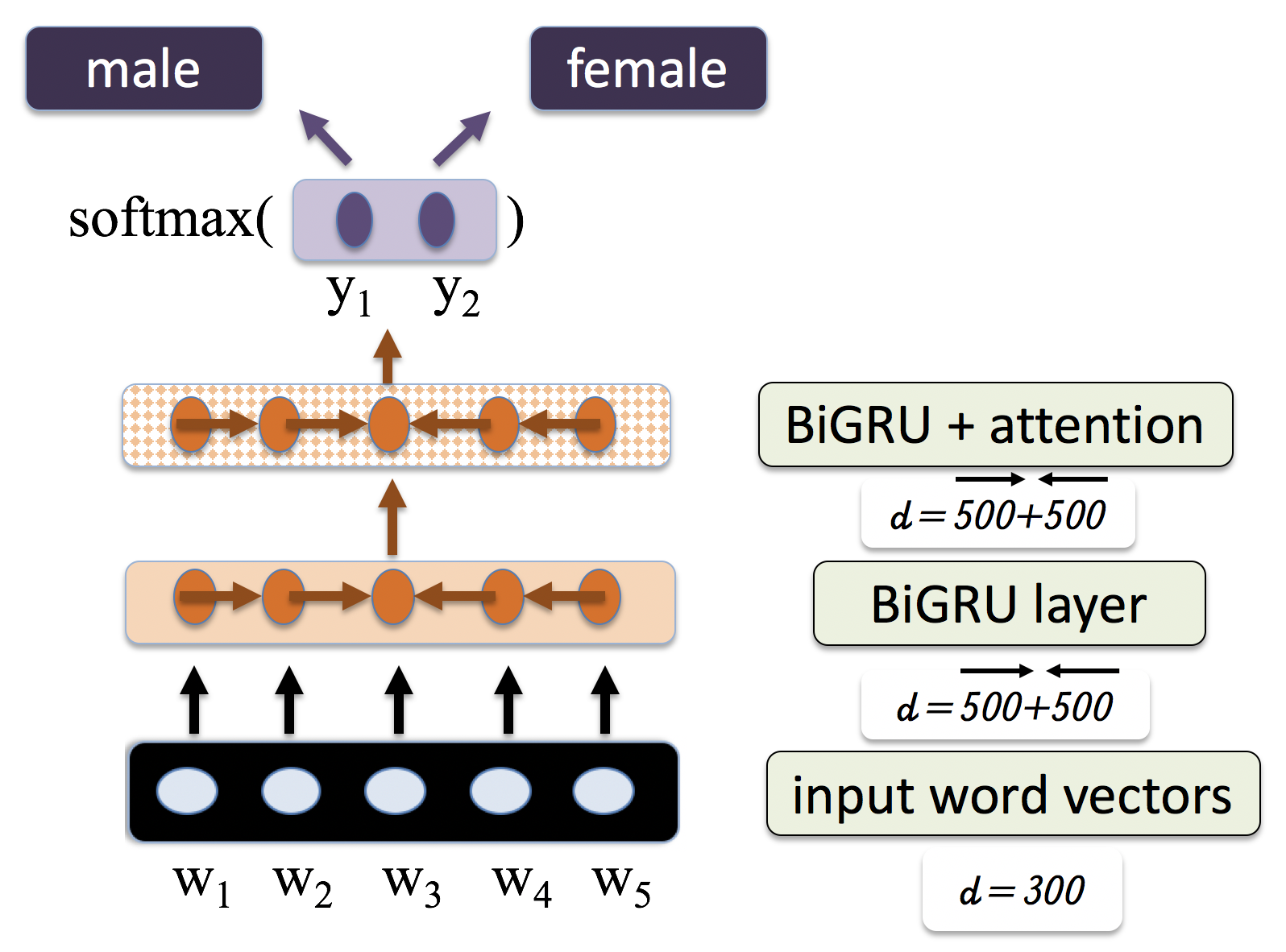}}
  \caption{Illustration of our network architecture for gender classification (single task network)}
  \label{fig:stl_net}
  \end{centering}
\end{figure}

\begin{figure*}[h]
\begin{centering}
 \frame{\includegraphics[width=\linewidth]{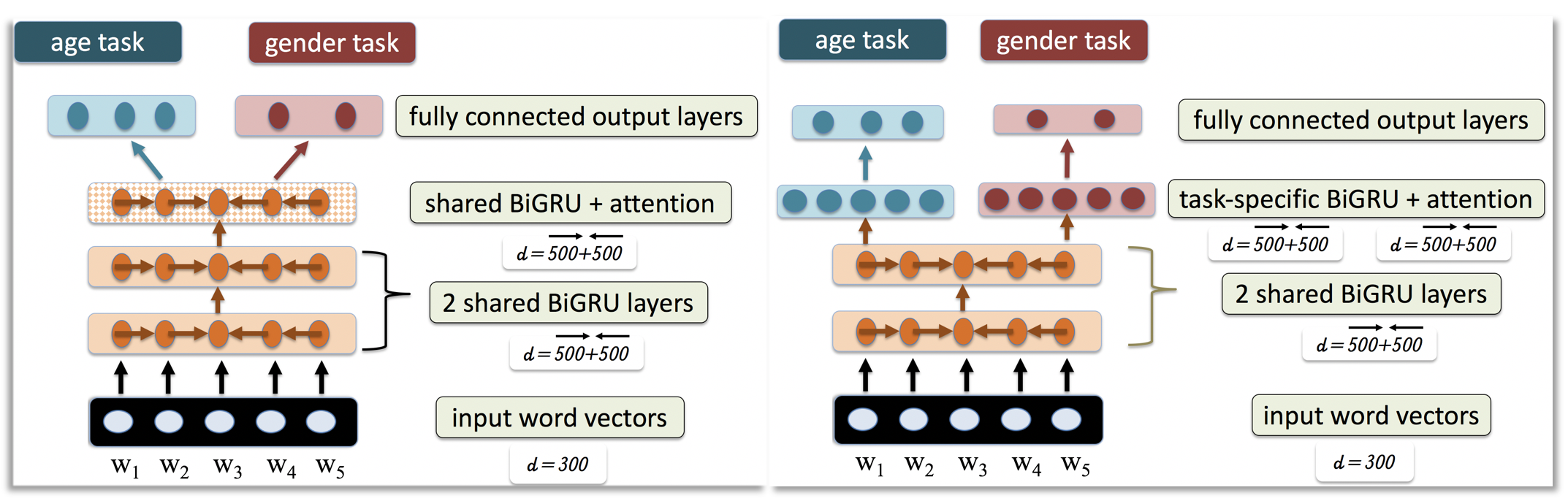}}
  \caption{Illustration of our 3-Layered MTL networks. \textbf{Left}: Network with shared attention; \textbf{Right}: Network with task-specific attention. We visualize 2 shared BiGRU layers, to emphasize the idea of sharing hidden layers. Our best performance, however, is acquired with only 1 shared BiGRU layer at the bottom of each network, followed by another BiGRU layer (either shared or task-specific) on which attention is applied.}
  \label{fig:mtl_nets}
  \end{centering}
\end{figure*}
We first design an independent network for each of the two tasks. We use the same network architecture across the 2 tasks. For each network, we experiment with 2 and 3 hidden BiGRU layers, each with 1,000 units (500 units in each direction).~\footnote{We found the networks with 3 hidden layers to overfit and so we report only networks with 2 hidden layers in all our experiments.} We add Transformer-style attention \textit{only} to the last hidden layer. We trim each sequence at 30 words, and use a batch size of 8. Each word in the input sequence is represented as a vector of 300 dimensions learned directly from the data. Word vectors weights $W$ are initialized with a standard normal distribution, with $\mu=0$, and $\sigma=1$, i.e., $W \sim N(0,1)$. Figure~\ref{fig:stl_net} offers an illustration of the single-task network for gender. The network for age is similar, with the exception that its output layer has the 3 units (for the 3 age classes in the data). For optimization, we use Adam~\cite{kingma2014adam} with a fixed learning rate of $1e-3$. For regularization, we use dropout~\cite{srivastava2014dropout} with a value of $0.5$ on each of the hidden layers. We run each network for 10 epochs, with early stopping, choosing the model that performs highest on DEV as our best model. We then run each best model on TEST, and report accuracy. Our best results on age are acquired with 3 epochs, and for gender with 2 epochs. Tables~\ref{tab:res_dev} and~\ref{tab:res_test} presents our results on DEV and TEST respectively. (The tables also show results for our subsequent experiments with MTL and BERT on DEV and TEST). 
On TEST, Attn-BiGRU is 10.78\% better than small-GRU for age and 8.30\% higher than the majority class baseline for gender. We note that results on DEV are very close to those on TEST. 
We now describe our MTL experiments.

\subsection{MTL BiGRUs for Age and Gender}\label{subsec:mtl}
\begin{table}[h]
\centering
\begin{tabular}{l|cc} \hline 
\textbf{Setting} & \textbf{Age} & \textbf{Gender} \\ \hline 
\textbf{maj-base} & 35.53 & 53.47 \\
\textbf{small-GRU} & 36.13 & 53.39 \\ \hline 
\textbf{Attn-BiGRU} & \underline{47.03} & \underline{61.64} \\
\hline 
\textbf{MTL-common-attn} & 47.85 & 62.50 \\ 
\textbf{MTL-spec-attn} & 47.92 & \underline{63.09} \\
\textbf{MTL-sprvsd-spec-attn} & \underline{48.23} & 62.99 \\\hline 
\textbf{MTL-MI-spec-attn} & 47.90 & \underline{63.15} \\ \hline

\textbf{BERT} & \textbf{50.95} & \textbf{65.31} \\ \hline 
\end{tabular}%
\caption{Model performance in accuracy on DEV. Baseline 1, maj-base, is the majority class in our TRAIN. Baseline 2, small-GRU is a unidirectional, 500-unit, one-layered GRU. For each block, best results that are higher than the baselines \textit{and} the best results in preceding block(s) are \underline{underlined}. Best result in each task is in \textbf{bold}. For single task models, best results for age are with 3 epochs and for gender with 2 epochs. For multi-task models, best accuracy is acquired with epochs between 3 and 6.~\footnote{We provide a table in the supplementary material with information about best epoch per model.}}
\label{tab:res_dev}
\end{table}

\begin{table}[h]
\centering

\begin{tabular}{l|cc} \hline 
\textbf{Setting} & \textbf{Age} & \textbf{Gender} \\ \hline 
\textbf{maj-base} & 35.53 & 53.47 \\
\textbf{small-GRU} & 36.29 & 53.37 \\ \hline 
\textbf{Attn-BiGRU} & \underline{47.07} & \underline{61.77} \\ \hline

\textbf{MTL-common-attn} & 47.82 & 62.36 \\ 
\textbf{MTL-spec-attn} & 47.99 & \underline{63.04} \\ 
\textbf{MTL-sprvsd-spec-attn} & \underline{48.25} & 62.81 \\ \hline 
\textbf{MTL-MI-spec-attn} &47.81 & \underline{63.13} \\ \hline
\textbf{BERT} & \textbf{51.42} & \textbf{65.30} \\ \hline 
\end{tabular}%
\caption{Model performance in accuracy on TEST.}
\label{tab:res_test}
\end{table}

To learn the two tasks simultaneously, we use a single network with 2 hidden layers each with the same dimensionality as the single-task networks (1,000 units in each layer).~\footnote{We also experiment with a network with 3 hidden layers, but we find it sub-optimal and so we do not discuss it further.} For comparison, we also use the same hyper-parameters as the single-task models for optimization with Adam, dropout regularization, and run for 10 epochs. We share the parameters of the first (bottom) hidden BiGRU layer across the three tasks. For the \textit{attention BiGRU layer} (2nd layer), we have \textit{three settings}:

\begin{table}[h]
\centering 
\begin{tabular}{|c|l|c|}
\hline
\textbf{Word}       & \textbf{Eng} & 
\textbf{MTL-S}    \\ \hline
<يسالونني>            &    they ask me     
& 0.169779 \\ \hline
<ياغالين>             &  you dear (+plural)       
& 0.164816 \\ \hline
<منيحاا>              &   beautiful (+fem)      
& 0.164619 \\ \hline
<مسساء>               &  evening        
& 0.161141 \\ \hline
<سيصادفك>             &   you'll meet by chance      
& 0.153956 \\ \hline
<بونسوار>             &  good evening (French)       
& 0.147900   \\ \hline
<سيصيبك>              &    it'll befall you      
& 0.147765 \\ \hline
<شلوونكم>             &   how are you (+plural)      
& 0.146776 \\ \hline
<امبيه>               &   I want it      
& 0.146680  \\ \hline
 <ياعيون>              &   you, darling       
 & 0.142256 \\ \hline
 <ياصدفه>              &     what a surprise     
 & 0.142011 \\ \hline
 <انزيين>              &    we beautify    
 & 0.141639 \\ \hline
 <صبحهم>               &  good morning        
 & 0.139647 \\ \hline
 <ياسااتر>             &  God protect      
 &  0.138652 \\ \hline
 <سابتسم>              &    I'll smile  
 &  0.138609 \\ \hline
\end{tabular}
\caption{Top 15 most highly weighted words based on average attention weights from our MTL-spec-attn (MTL-S) network for \textit{gender}.}
\label{tab:word_attention}
\end{table}

\textbf{(i) Shared Attention:}
In this setting, we share the attention BiGRU layer across the two tasks. As such, this network has all its layers shared across the two tasks. We refer to this network as \textit{MTL-common-attn}. This architecture is aimed at investigating the utility of allowing the network to attend simultaneously to each of the two tasks. 

\textbf{(ii) Task-Specific Attention:}
For this network, while the first hidden layer(s) are shared, the last layer (attention layer) is task-specific (i.e., independent for each task). We will call this model \textit{MTL-spec-attn}. The goal of this architecture is to investigate the feasibility of allowing each task to specialize its own attention, while also sharing information (i.e., the bottom layer) with the other task.

\textbf{(iii) \textit{Supervised} Task-Specific Attention:} We perform supervised attention based on mutual information (MI) with a list of 500 weighted words, as explained in Section~\ref{subsec:suprvsdattn}. 
As Table~\ref{tab:res_test} shows, for the two tasks, MTL is consistently better than single task learning (and our baselines). We observe that task-specific attention is superior to shared attention across the two tasks, in both DEV and TEST. We also observe that supervised attention improves over regular self-attention for the age task (0.31\% acc on DEV and 0.26\% on TEST), but not for gender. This shows that while the MTL is useful, it is better to specialize the attention layer. It seems this specialized architecture allows the network to identify representations that are harder to disentangle in the shared attention setting.

\textbf{\textit{MI-Based MTL:}} Rather than using frequency as a basis for choosing our input vocabulary, we experiment with using a vocabulary weighted by MI. More specifically, we use the top 100K MI-weighted words as our vocabulary in the embedding layer. For this setting, we only run our best MTL network from previous experiments (MTL with task-specific attention). We refer to this network as MTL-MI-spec-attn. We note that this network improves performance slightly over all our previous models on the gender task. 

\subsubsection{Interpreting Attention in MTL}
We have established the effectiveness of attention in the context of our MTL networks. To further understand model behavior, we extract words most highly attended to by our MTL-spec-attn network for the gender task. Table~\ref{tab:word_attention} provides the top 15 examples among these words. Intuitively, these words seem relevant for distinguishing gender. We provide more samples of words attended to by our single and MTL models in the supplementary material, along with corresponding MI scores. We observe these words are assigned very low MI scores and are not top ranked by MI~\footnote{MI scores and neural model weights are not directly comparable, but we find words with higher weights from the MTL-spec-attn network to be more intuitive}, thereby highlighting how the neural network is capable of capturing nuances not necessarily identified by the statistical approach. We hypothesize this is partially because MI does not operate over distributed representations of words, but rather treat them as atomic symbols.~\footnote{We provide more samples of words attended to by our single and MTL models in the supplementary material, along with corresponding MI scores.} For further probing into our model behavior, we provide an attention-based visualization of a number of hand-picked samples along with their English translations in the supplementary material.

\subsection{BERT-Based Modeling}\label{subsec:bert_results}
We use the BERT-Base, Multilingual Cased model released by the authors~\footnote{\url{https://github.com/google-research/bert/blob/master/multilingual.md}.} to model each of our two tasks. The model is trained on 104 languages, including Arabic, with 12 layer, 768 hidden units each, 12 attention heads, and has 110M parameters. The model has 119,547 word pieces shared across all the languages. For fine-tuning, we use a maximum sequence size of 30 words and a batch size of 32. We set the learning rate to 2e-5 and train for 15 epochs. From Tables~\ref{tab:res_dev} and~\ref{tab:res_test}, we can see that BERT performs consistently better than the two baselines on the two tasks, irrespective of the data split. 
On TEST, it improves with 15.13\% acc. (for age) and 11.83\% acc. (for gender) over the best baseline. BERT also outperforms our best MTL models. 
Also, BERT outperforms MTL-suprvsd-spec-attn on age with 3.17\% acc and MTL-spec-attn on gender with 2.17\% acc on TEST. These results demonstrate (i) the effectiveness of BERT on our two tasks, and (ii) the relative utility of our MTL models, which still perform well even though they do not need pre-training on a huge amount of data like BERT.

\section{Model Evaluation on External Data}\label{sec:external_eval}
\textbf{Tweet-Level Gender Evaluation:} In order to test how our best model (BERT) will generalize to external data, we develop a new, additional dataset \texttt{EXTERNAL} only for the gender task. To acquire EXTERNAL, we manually label an in-house dataset of $\sim$ 2,124 users from 20 Arab countries (including all countries in Arab-Tweet)~\footnote{Users are manually assigned country tags in the data.} with gender tags. EXTERNAL has data from 550 \textit{female} users, 1,335 \textit{male} users, and 239 users with \textit{unknown} gender. For a balanced dataset, we only keep each gender at 550 users, for a total of 1,100 users contributing $\sim$1,320K tweets. We split EXTERNAL into DEV-EXT and TEST-EXT. We run BERT on each of these splits. We acquire 61.31\% acc on DEV-EXT and 61.46\% acc on TEST-EXT. We note these results are 4\% and 3.84\% acc \textit{less} than our performance on Arab-Tweet DEV and TEST, respectively. We hypothesize this could be due to (i) diachronic degradation of language models~\citep{jaidka2018diachronic}, since EXTERNAL covers only up until the year 2017; but also (ii) EXTERNAL being harder and more diverse since, unlike Arab-Tweet, it was not collected with a seed word approach.  

\section{Related Works}\label{sec:rel}

\subsection{Arabic}
Arabic is a collection of language varieties, including a standard variety that is referred to as modern standard Arabic (MSA). MSA is more popular in pan-Arabic media and formal communication (such as in educational settings) and literary and religious venues. Dialectal Arabic is used widely by Arabs in their daily lives. 
Arabic varieties differ at various linguistic levels, including phonological, morphological, lexical, and syntactic differences~\cite{holes2004modern,bassiouney2009arabic,palva2006dialects}. Traditionally, Arabic dialects have been categorized based on coarse-grained geographical divisions that cross country boundaries, e.g., \cite{habash2010introduction,versteegh2014arabic}. More recently, there has been interest in more fine-grained differences, e.g.,~\cite{mubarak2014using,sadat2014automatic,mageed2018city,salameh2018fine,qwaider2018shami}. 

\subsection{Author Profiling}
Detecting sociological and psychological attributes of a user is usually referred to as \textit{author profiling}. This typically involves tasks such as detecting author gender and age (see Section~\ref{subsec:age_gender}), personality~\citep{schwartz2013personality,matz2017psychological,bleidorn2018using,hinds2019human}, political orientation~\citep{colleoni2014echo,preoctiuc2017beyond}, and moral traits~\citep{johnson2018classification,pang2019language}. Related terms include \textit{author attribution} and \textit{forensic authorship analysis}~\citep{wright2013using}, where efforts are focused on identifying the specific author of a given text. Text-based approaches to author profiling (e.g., ~\citep{gamon2004linguistic,argamon2019register}) are rooted in computational stylometry~\citep{goswami2009stylometric,verhoeven2018two} and can be traced back to the work of~\citet{holmes1998evolution}. From a linguistic perspective, this line of work is based on the concept of \textit{idiolect}, the tenet that language is a property of the individual~\cite{hudson1996sociolinguistics,barlow2013individual,barlow2018individual} as contrasted to the idea of language being the property of a speech community~\citep{labov1989exact}. Network-based approaches involve making use of online relations such as friendships, followings, mentions, and comments~\citep{mitrou2014social} for author profiling. Notably, the PAN author profiling shared task~\cite{rangel2013overview,rangel2014overview,rangel2015overview,rangel2016overview,rangel2017overview} was launched to progress work in this area. A summary of the PAN work can be found in~\citep{potthast2019decade}

\subsection{Age and Gender}\label{subsec:age_gender} 
Several works have been conducted on English-based age and gender detection. These include~\cite{rao2010classifying,flekova2016exploring,burger2011discriminating,verhoeven2016twisty,daneshvar2018gender}. Several of these works make use of text n-gram and topic models features~\citep{schwartz2013personality,sap2014developing}. In addition, for age, many studies cast the as multi-class classification, with several age group categories (e.g., ages 10-19, 20-29, 30-39). Other works cast age as a regression~\citep{nguyen2011author} or it with classification~\cite{chen2019joint}. There are also works that target age and gender prediction from facial data~\citep{levi2015age} or a mixture of text and image data~\citep{vijayaraghavan2017twitter}. We also note that most of the literature is on English, with only few focusing on cross-lingual settings for gender ~\citep{van2018bleaching}, and age and personality~\citep{verhoeven2016twisty}. Minimal works exist on other languages~\citep{verhoeven2018two}. For Arabic, with the exception of some recent treatments~\cite{rangel2017overview,alrifai2017arabic}, there is not much work we know of that is focused on learning these tasks.

\section{Conclusion}\label{sec:conc}
In this work, we reported our models targeting age and gender detection in Arabic social media. We have shown the utility of MTL, especially with task-specific attention, on the two learning tasks. Even though our models do not need to be pre-trained on huge amounts of data, as is the case with BERT, they perform well on both tasks. We have also shown the utility of the newly-released BERT model. Our models can be deployed for real-world social media analysis at scale, since they can cover 17 Arabic countries. The models are also language-agnostic, and so can be applied to other languages. This is one of our future directions.

\bibliography{acl2020.bib}
\bibliographystyle{acl_natbib}
\newpage
\appendix

\section{Summary of Supplementary Material}\label{sec:summary}
\begin{figure*}[h]
\begin{centering}
\frame{\includegraphics[width=\linewidth]{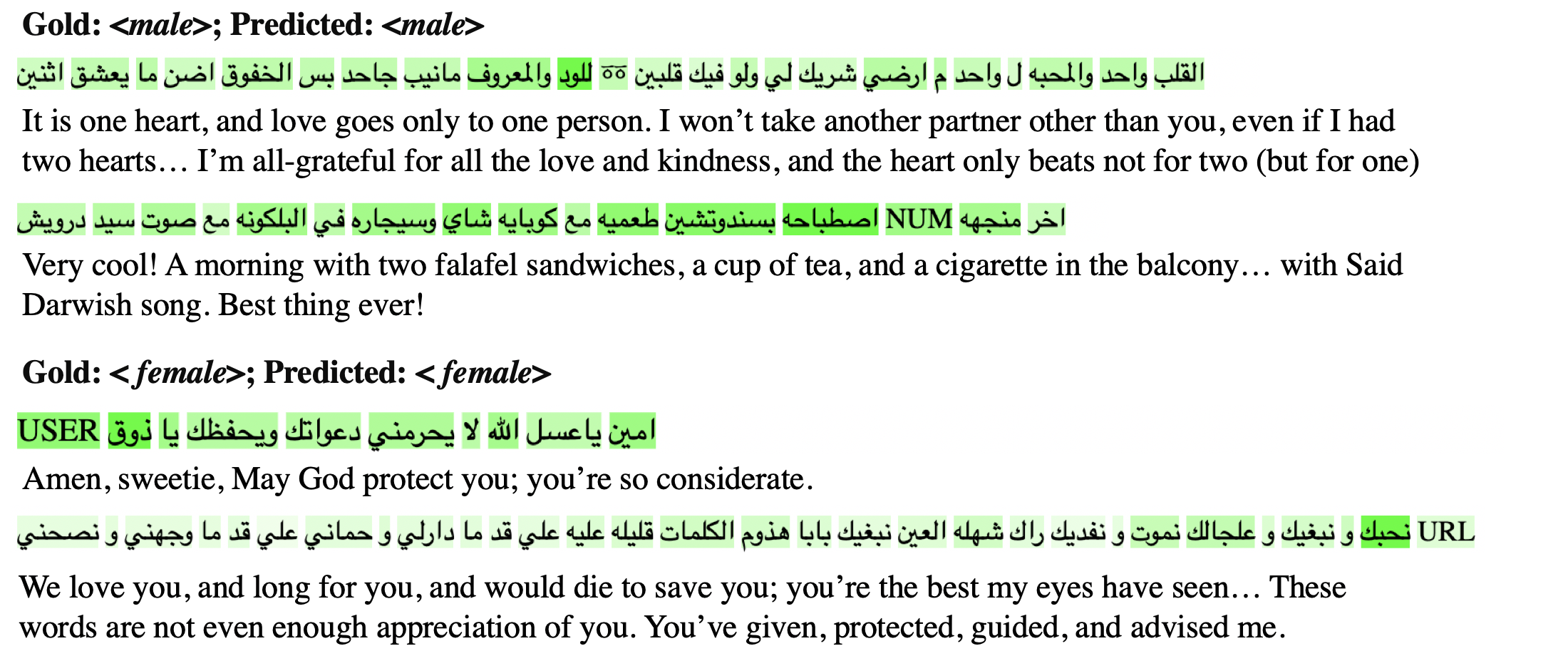}}
  \caption{Attention-based visualization based on our MTL-spec-attn of hand-picked, correctly predicted samples from our TRAIN data.}
  \label{fig:attn_viz_correct}
  \end{centering}
\end{figure*}

\begin{figure*}[h]
\begin{centering}
\frame{\includegraphics[width=\linewidth]{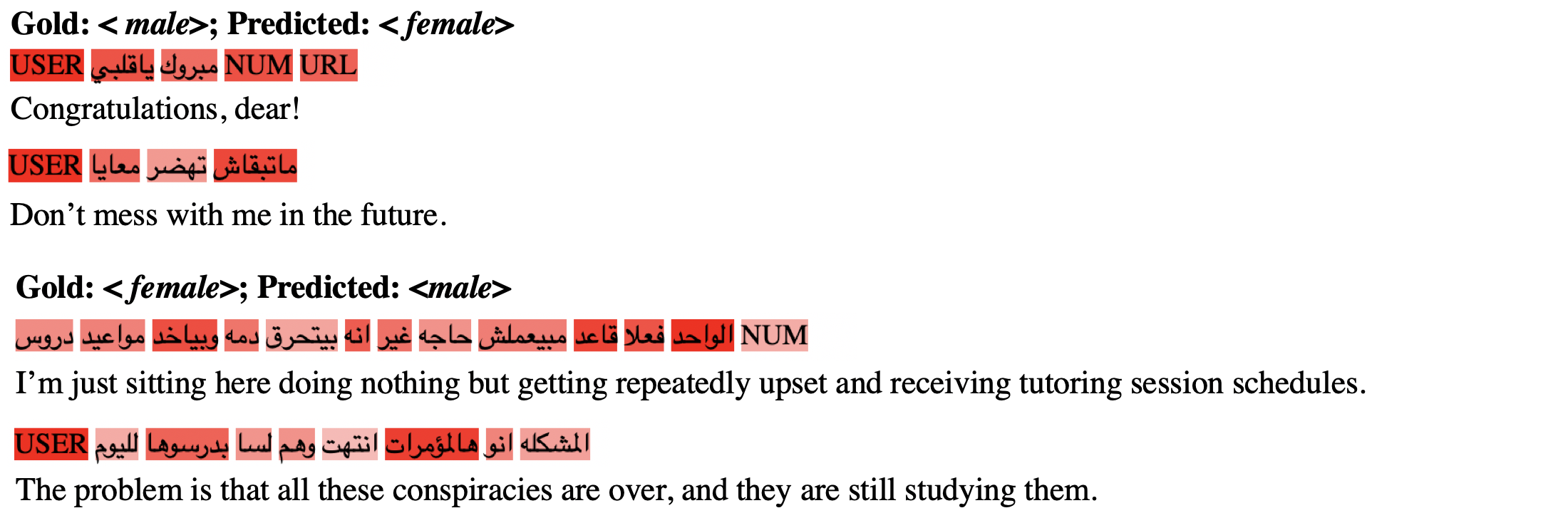}}
  \caption{Attention-based visualization based on our MTL-spec-attn of hand-picked, incorrectly predicted samples from our TRAIN data.}
 \label{fig:attn_viz_incorrect}
  \end{centering}
\end{figure*}

We provide the following supplementary items:
\begin{enumerate}
   \item Table~\ref{tab:topMTL} shows top 20 words attended to by our MTL-spec-attn model, along with attention weights by other (MTL) models and MI.
    \item Table~\ref{tab:topMI} shows top 20 words based on MI scores, along with attention weights by other models.
    \item Figures~\ref{fig:attn_viz_correct} and ~\ref{fig:attn_viz_incorrect} provides examples of attention-based visualization of our MTL-spec-attn model on the gender task.
      \item Table~\ref{tab:model_epoch} provides information about the epoch acquiring best performance across all our models on DEV.
\end{enumerate}

\begin{table*}[]
\centering 
\begin{tabular}{|c|l|cccc|}
\hline
\textbf{Word}       & \textbf{Eng} & \textbf{MI} & \textbf{Attn-BiGRU} & \textbf{MTL-C} & \textbf{MTL-S}    \\ \hline
<يسالونني>            &    they ask me       & 1.00E-06    & 0.038055            & 0.119423       & 0.169779 \\ \hline
<ياغالين>             &  you dear (+plural)         & 9.40E-05    & 0.052512            & 0.078695       & 0.164816 \\ \hline
<منيحاا>              &   beautiful (+fem)        & 5.00E-06    & 0.031567            & 0.117903       & 0.164619 \\ \hline
<مسساء>               &  evening         & 4.00E-06    & 0.027612            & 0.142467       & 0.161141 \\ \hline
<سيصادفك>             &   you'll meet by chance        & 0.00E+00    & 0.040002            & 0.103273       & 0.153956 \\ \hline
<بونسوار>             &  good evening (French)         & 3.00E-05    & 0.044578            & 0.10257        & 0.1479   \\ \hline
<سيصيبك>              &    it'll befall you       & 2.00E-06    & 0.031588            & 0.097372       & 0.147765 \\ \hline
<شلوونكم>             &   how are you (+plural)        & 5.00E-06    & 0.0352              & 0.125699       & 0.146776 \\ \hline
<امبيه>               &   I want it        & 1.50E-05    & 0.041253            & 0.095131       & 0.14668  \\ \hline
<ياعيون>              &   you, darling        & 0.00E+00    & 0.032135            & 0.123004       & 0.142256 \\ \hline
<ياصدفه>              &     what a surprise      & 1.00E-06    & 0.025398            & 0.124698       & 0.142011 \\ \hline
<انزيين>              &    we beautify       & 1.00E-06    & 0.025684            & 0.096841       & 0.141639 \\ \hline
<صبحهم>               &  good morning         & 0.00E+00    & 0.03443             & 0.118689       & 0.139647 \\ \hline
<ياسااتر>             &  God protect       & 0.00E+00    & 0.036266            & 0.104663       & 0.138652 \\ \hline
<سابتسم>              &    I'll smile       & 1.00E-06    & 0.029951            & 0.097428       & 0.138609 \\ \hline
<تقييمكم>             &  what's your evaluation         & 2.00E-06    & 0.034339            & 0.091574       & 0.137949 \\ \hline
<مسلخير>              &      good evening     & 7.00E-06    & 0.036409            & 0.114726       & 0.137911 \\ \hline
<االخيرر>             &    good (morning/evening)       & 0.000141    & 0.058663            & 0.094799       & 0.137106 \\ \hline
<ليلهه>               &    night       & 9.00E-06    & 0.026944            & 0.104899       & 0.136817 \\ \hline
<نبئ> &    we want       & 0.00E+00    & 0.036886            & 0.098703       & 0.136785 \\ \hline
\end{tabular}
\caption{Top 20 most highly weighted words based on average attention weights from our MTL-spec-attn (MTL-S) network for \textit{gender}. We also add corresponding average attention weights allocated to the words by our single-task attention BiGRU model (Attn-BiGRU) and MTL-common-attn (MTL-C) model. In addition, we add the mutual information (MI) weights for these words. Note the MI values are very small. }
\label{tab:topMTL}
\end{table*}

\begin{table*}[]
\centering 
\begin{tabular}{|c|l|cccc|}
\hline
\multicolumn{1}{|c|}{\textbf{Word}} & \textbf{Eng} & \textbf{MI} & \textbf{Attn-BiGRU} & \textbf{MTL-C} & \textbf{MTL-S} \\ \hline
\multicolumn{1}{|c|}{<حبيبتي>}        &my darling (+fem.) & 0.002155                   & 0.050888             & 0.077016     & 0.094471     \\ \hline
\multicolumn{1}{|c|}{<في>}          & in     & 0.001078                   & 0.033693             & 0.047777     & 0.047659     \\ \hline
                <يارب>           & Oh, Lord      & 0.001053                   & 0.035691             & 0.069558     & 0.0758       \\ \hline
                <عارفه>          & I know (+fem.)      & 0.000664                   & 0.043761             & 0.06062      & 0.068043     \\ \hline
                                <انا>   & I & 0.000559                   & 0.034185             & 0.066936     & 0.072569     \\ \hline
                               <قلبي>   & my heart & 0.00047                    & 0.03583              & 0.058327     & 0.060456     \\ \hline
<الدوري>              &     soccer season            & 0.000457                   & 0.039459             & 0.049224     & 0.051559     \\ \hline
<القصاص>             & retributive justice                  & 0.000444                   & 0.052651             & 0.046728     & 0.068582     \\ \hline
<مباراه>               & game                & 0.000434                   & 0.037312             & 0.057625     & 0.060602     \\ \hline
<بن>                   & son of                & 0.000431                   & 0.034937             & 0.04922      & 0.048669     \\ \hline
<لاعب>                  & player               & 0.000407                   & 0.038375             & 0.055207     & 0.055142     \\ \hline
<انتي>                   & you (+fem.)              & 0.000399                   & 0.034844             & 0.065708     & 0.068472     \\ \hline
<رئيس>                  & president               & 0.00039                    & 0.034811             & 0.057263     & 0.059484     \\ \hline
<علي>                    & Ali/on              & 0.000382                   & 0.033676             & 0.049267     & 0.04817      \\ \hline
<يالغالي>                 & dear (+you)             & 0.000358                   & 0.043446             & 0.069519     & 0.072201     \\ \hline
<ميسي>                    & Messi             & 0.000325                   & 0.038833             & 0.061201     & 0.067188     \\ \hline
<الهلال>                  & Hilal (soccer team)             & 0.000319                   & 0.037964             & 0.053231     & 0.057722     \\ \hline
<يابو>                   & father of (+you)              & 0.000302                   & 0.040432             & 0.064222     & 0.070818     \\ \hline
<برشلونه>                 & Barcelona             & 0.000299                   & 0.03753              & 0.055462     & 0.060405     \\ \hline
<كاس>                     & cup (soccer)             & 0.000279                   & 0.036868             & 0.04998      & 0.048994     \\ \hline
\end{tabular}
\caption{Top 20 highest weighted words based on mutual information (MI) for \textit{gender}. We also add corresponding average attention weights allocated to the words by our single-task attention BiGRU model (Attn-BiGRU), MTL-common-attn (MTL-C), and MTL-spec-attn (MTL-S) networks.}
\label{tab:topMI}

\end{table*}

\begin{table*}[h]
\centering 
\begin{tabular}{l|cc|cc}
\hline
\textbf{Setting}              & \textbf{Age} & \multicolumn{1}{l|}{\textbf{Epoch}} & \textbf{Gender} & \multicolumn{1}{l}{\textbf{Epoch}} \\ \hline
\textbf{small-GRU}            & 36.13        & 1                                      & 53.39           & 1                                     \\ \hline
\textbf{Attn-BiGRU}           & 47.03        & 3                                        & 61.64           & 2                                       \\
\textbf{MTL-common-attn}      & 47.85        & 6                                        & 62.50           & 4                                       \\
\textbf{MTL-spec-attn}        & 47.92        & 4                                        & 63.09           & 4                                       \\
\textbf{MTL-sprvsd-spec-attn} & 48.23        & 5                                        & 62.99           & 3                                       \\ \hline
\textbf{MTL-MI-spec-attn}     & 47.90        & 3                                        & 63.15           & 5                                       \\ \hline
\textbf{BERT}                 & 50.95        & 4                                        & 64.96           & 6                                       \\ \hline
\end{tabular}
\caption{Epoch information for best model performance on our DEV.}
\label{tab:model_epoch}
\end{table*}

\end{document}